\documentclass[twocolumn,10pt]{asme2ej}

\usepackage{graphicx} %% for loading jpg figures
\usepackage{hyperref}   % to set up hyperlinks

\usepackage{amsmath, amssymb}%, amsthm}
\usepackage{cite, multirow}
\usepackage[caption = false, font = footnotesize]{subfig}
\hypersetup{
	colorlinks=true,
	linkcolor=blue,
	citecolor=blue,
	urlcolor=blue,
}
\title{XTENTH-CAR: A Proportionally Scaled Experimental Vehicle Platform for Connected Autonomy and All-Terrain Research}

\author{Shathushan Sivashangaran \footnote[1]{\thanks{Corresponding author.\\
XTENTH-CAR source code, hardware and software installation guides available at: \href{https://github.com/Shathushan-Sivashangaran/XTENTH-CAR}{https://github.com/Shathushan-Sivashangaran/XTENTH-CAR}\\
© 2023 ASME. This work has been accepted to ASME for publication.}}\\
    \affiliation{ Doctoral Researcher\\
	Autonomous Systems and Intelligent Machines\\ 
    Laboratory\\
	Department of Mechanical Engineering\\
	Virginia Tech\\
	Blacksburg, VA 24061, USA\\
    Email: shathushansiva@vt.edu
    }	
}

\author{Azim Eskandarian \\
    \affiliation{Professor and Department Head\\
	Autonomous Systems and Intelligent Machines\\
    Laboratory\\
	Department of Mechanical Engineering\\
	Virginia Tech\\
	Blacksburg, VA 24061, USA\\
    Email: eskandarian@vt.edu
    }
}

\begin{document}

\maketitle    
%%%%%%%%%%%%%%%%%%%%%%%%%%%%%%%%%%%%%%%%%%%%%%%%%%%%%%%%%%%%%%%%%%%%%%

\begin{abstract}
{\it Connected Autonomous Vehicles (CAVs) are key components of the Intelligent Transportation System (ITS), and all-terrain Autonomous Ground Vehicles (AGVs) are indispensable tools for a wide range of applications such as disaster response, automated mining, agriculture, military operations, search and rescue missions, and planetary exploration. Experimental validation is a requisite for CAV and AGV research, but requires a large, safe experimental environment when using full-size vehicles which is time-consuming and expensive. To address these challenges, we developed XTENTH-CAR (eXperimental one-TENTH scaled vehicle platform for Connected autonomy and All-terrain Research), an open-source, cost-effective proportionally one-tenth scaled experimental vehicle platform governed by the same physics as a full-size on-road vehicle. XTENTH-CAR is equipped with the best-in-class NVIDIA Jetson AGX Orin System on Module (SOM), stereo camera, 2D LiDAR and open-source Electronic Speed Controller (ESC) with drivers written for both versions of the Robot Operating System (ROS 1 \& ROS 2) to facilitate experimental CAV and AGV perception, motion planning and control research, that incorporate state-of-the-art computationally expensive algorithms such as Deep Reinforcement Learning (DRL). XTENTH-CAR is designed for compact experimental environments, and aims to increase the accessibility of experimental CAV and AGV research with low upfront costs, and complete Autonomous Vehicle (AV) hardware and software architectures similar to the full-sized X-CAR experimental vehicle platform, enabling efficient cross-platform development between small-scale and full-scale vehicles.
}
\end{abstract}

%%%%%%%%%%%%%%%%%%%%%%%%%%%%%%%%%%%%%%%%%%%%%%%%%%%%%%%%%%%%%%%%%%%%%%
%\begin{nomenclature}
%\entry{A}{You may include nomenclature here.}
%\entry{$\alpha$}{There are two arguments for each entry of the nomemclature environment, the symbol and the definition.}
%\end{nomenclature}

%The primary text heading is  boldface and flushed left with the left margin.  The spacing between the  text and the heading is two line spaces.

%%%%%%%%%%%%%%%%%%%%%%%%%%%%%%%%%%%%%%%%%%%%%%%%%%%%%%%%%%%%%%%%%%%%%%
\section{Introduction}

Connected Autonomous Vehicles (CAVs) are integral to the future of transportation, and have great potential to improve traffic flow, safety and fuel efficiency. CAVs are Autonomous Vehicles (AVs) with vehicle connectivity enabled for vehicle-to-vehicle (V2V), vehicle-to-infrastructure (V2I) and vehicle-to-everything (V2X) communications \cite{eskandarian2019research, duarte2018impact}. CAVs improve upon AV benefits to transportation via acquisition of information beyond the ego vehicle's Field of View (FoV) to achieve superior traffic throughput and energy economy. 

All-terrain Autonomous Ground Vehicles (AGVs) are beneficial tools for a wide range of applications, such as disaster response, automated mining, agriculture, military operations, search and rescue missions, and planetary exploration \cite{victerpaul2017path}. Intelligent navigation algorithms that incorporate Deep Reinforcement Learning (DRL) enable AGVs to cognitively chart collision-free motion trajectories in environments without a-priori maps \cite{sivashangaran2023deep, sivashangaran2021intelligent}, which is a necessity in information poor, dynamically altering environments.

AGVs and CAVs are a continuously evolving research domain with substantial developments over the past decade. The complex nature of real-world traffic environments, and safety considerations pose challenges to experimental research and expeditious advancement of CAV technology. Full-scale experimental vehicles integrated with the CARMA$^{\text{SM}}$ (Cooperative Automation Research Mobility Applications) platform developed by the U.S. Department of Transportation (USDOT) Federal Highway Administration (FHWA) \cite{carma2021products} that utilize affordable, high quality hardware such as X-CAR \cite{mehr2022XCAR} facilitate Cooperative Driving Autonomy (CDA) research and development. However, research that necessitates multiple vehicles requires a large, safe experimental environment, and becomes time-consuming and expensive using full-sized vehicles. Moreover, contemporary AGV DRL research is primarily driven by simulation experiments due to the high computation cost required for training and implementation of agents in the real-world.

To address these challenges, we developed XTENTH-CAR (eXperimental one-TENTH scaled vehicle platform for Connected autonomy and All-terrain Research), an open-source, cost-effective proportionally one-tenth scaled experimental vehicle platform with best-in-class embedded computing, governed by the same physics as a full-size on-road vehicle.

Proportionally scaled AV testbeds are pertinent to AV and CAV research pipelines, enabling rapid experimental validation of algorithms prior to implementation and testing in full-size vehicles. XTENTH-CAR shares similar hardware and software architectures to the full-size X-CAR platform, enabling cross-platform development and real-world evaluation of novel cooperative perception, localization and motion planning algorithms. Furthermore, XTENTH-CAR is equipped with a state-of-the-art embedded CPU and GPU unit to facilitate experimental AGV Artificial Intelligence (AI) research. The NVIDIA Jetson AGX Orin System on Module (SOM) on XTENTH-CAR is NVIDIA's most capable, newly released SOM, featuring up to eight times greater performance than the previous generation \cite{nvidia}. It contains 2048-core NVIDIA Ampere architecture GPU with 64 Tensor Cores, 12-core Arm 64-bit CPU and 32GB LPDDR5 RAM in a small footprint with high-speed interfaces to support DRL research in a real-world lab environment.

\begin{figure}[ht]
    \centering
    \includegraphics[width = \columnwidth]{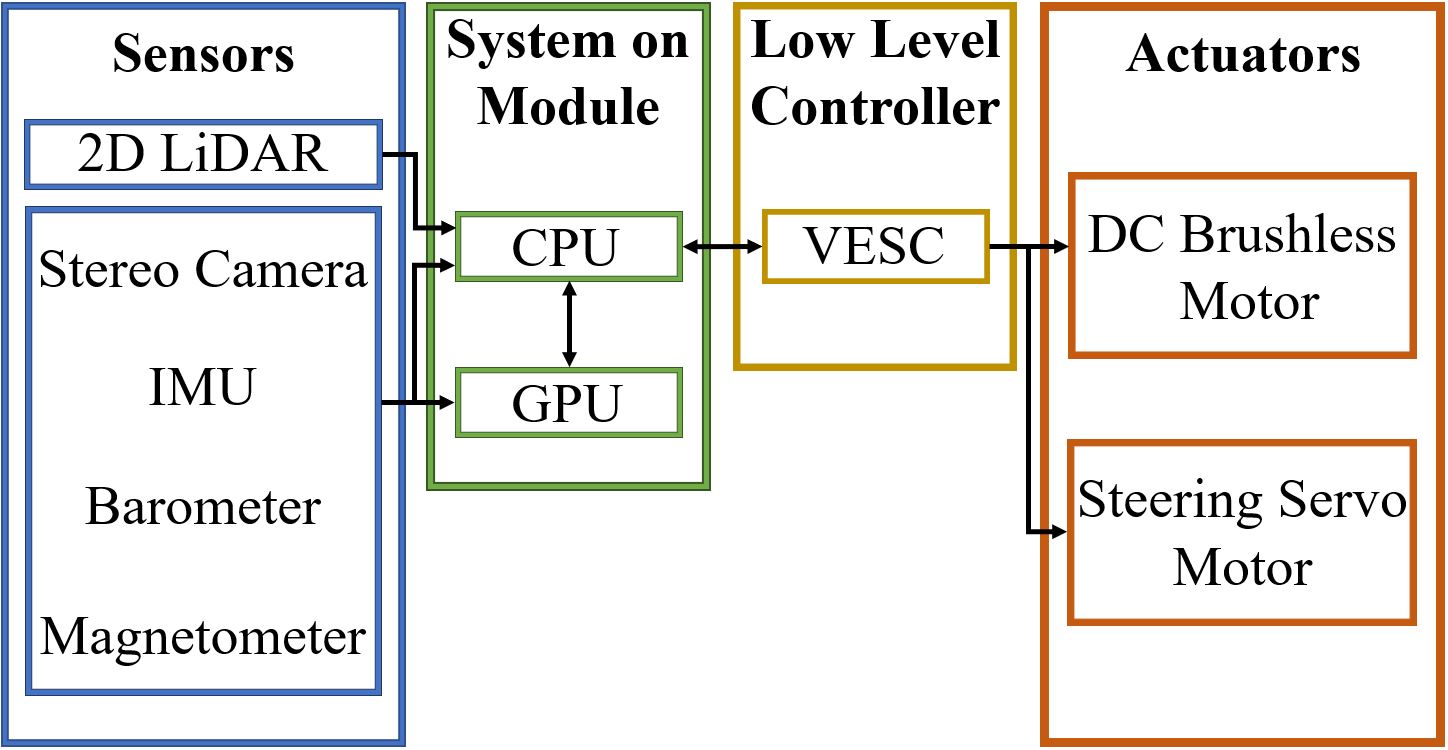}
    \setlength{\abovecaptionskip}{-\baselineskip}
    \caption{XTENTH-CAR hardware architecture and data flow.} \label{hwArch}
\end{figure}

The sensor suite comprises YDLIDAR G2, a 2D planar LiDAR, ZED 2 stereo camera with built-in Inertial Measurement Unit (IMU), barometer and magnetometer, and Vedder Electronic Speed Controller (VESC), an open-source Electronic Speed Controller (ESC) with drivers written for both versions of the Robot Operating System (ROS 1 \& ROS 2) to facilitate static and dynamic object detection, mapping, localization and odometry, enabling full CAV and AGV capabilities. XTENTH-CAR's hardware architecture, and data flow between components, are illustrated in Figure \ref{hwArch}.
 
This paper details XTENTH-CAR's chassis, actuators, computation unit, sensor suite and real-world performance. Simultaneous Localization and Mapping (SLAM) is implemented using Hector SLAM \cite{KohlbrecherMeyerStrykKlingaufFlexibleSlamSystem2011}, a technique that uses the Extended Kalman Filter (EKF) for positioning and Gauss-Newton method for mapping. Hector SLAM was chosen for its optional odometry requirement, and low error rates and computation requirements \cite{dhaoui2022mobile}. Mask R-CNN \cite{he2017mask}, and the Common Objects in Context (COCO)
dataset are utilized to validate Computer Vision (CV) and Machine Learning (ML) based object detection, tracking and identification. The CPU and memory usage for perception, control, and motion planning techniques that include collision avoidance using Model Predictive Control (MPC) integrated with Artificial Potential Function (APF) \cite{shang2023mpcapf}, and cognitive exploration utilizing DRL \cite{sivashangaran2023deep} are evaluated, and shown to execute efficiently in real-time. 

The rest of the paper is organized as follows: Section \ref{background} provides background on past proportionally scaled experimental AVs and CAVs, and the ROS middleware utilized in XTENTH-CAR for communication between sensors and actuators. Section \ref{actuation} outlines the actuators and chassis used to achieve accurate scaled on-road vehicle dynamics. XTENTH-CAR's computation and data processing SOM is discussed in Section \ref{computation}. The sensor suite utilized to enable a full range of on-road CAV and off-road AGV capabilities is detailed in Section \ref{sensing}. Section \ref{results} presents real-world environmental perception, actuator control and computation performance results, and finally, Section \ref{conclusions} concludes the paper. 

\section{Background} \label{background}

In this section, we review past proportionally scaled experimental AV platforms, and ROS.

\subsection{Related Work}

Early proportionally scaled experimental vehicle platforms utilized off-board computation, where sensor data is transmitted to a separate system for processing, following which control inputs are transmitted back to the vehicle. ETH Zurich built a 1/43 scale Ackermann steered AV platform using this mechanism for experimental evaluation of optimization based autonomous racing \cite{liniger2015optimization}.

Advancements in computer technology enabled real-time onboard computation in larger 1/10 proportionally scaled AVs. The Berkeley Autonomous Race Car (BARC) developed by University of California, Berkeley utilized an ODROID single-board computer for data processing, and custom wheel encoders to estimate pose and velocity, for autonomous drifting and Learning Model Predictive Control (LMPC) based autonomous racing \cite{gonzales2016autonomous, rosolia2019learning}. Limitations of this platform include low computing capability of the ODROID, and dependence on wheel encoders built using light sensors.

The launch of NVIDIA's Jetson series SOM with embedded CPU and GPU, specifically designed to facilitate robotics research, aided implementation and evaluation of Machine Learning (ML) AV algorithms in scaled platforms. The MIT RACECAR utilized the NVIDIA Jetson TX1 and VESC for accurate odometry coupled with a sensor suite that included 2D LiDAR, stereo camera and IMU for autonomous racing and off-road applications \cite{mitracecar2017}.

Our lab at Virginia Tech developed ASIMcar \cite{asimcar2019}, a low-cost scaled CAV platform that utilized the NVIDIA Jetson TX2, VESC for actuator control, and cost effective sensor suite that included an IR proximity sensor, IMU and camera with fisheye lens. XTENTH-CAR is a dual purpose evolution of ASIMcar with a more durable chassis for off-road experiments, and cross-platform architecture compatibility with the full-scale X-CAR CAV platform. 

MuSHR developed by University of Washington utilizes the NVIDIA Jetson Nano which is now outdated, and no longer supported \cite{mushr2019}. The open-source F1TENTH platform developed by University of Pennsylvania utilizes the NVIDIA Jetson Xavier NX Developer Kit, and provides a testing-oriented simulator \cite{f1tenth2019}. This platform is primarily focused on robotics education and racing. The Xavier NX is lightweight, but not ideal for computationally expensive applications such as AGV DRL research. XTENTH-CAR is specifically built with best-in-class computation to facilitate state-of-the-art experimental CAV and AGV research.

\subsection{Robot Operating System}

ROS is an open-source modular framework and set of tools for robot software development \cite{quigley2009ros, quigley2015programming} that provides OS functionality on a heterogeneous computer cluster. A process on the loosely coupled ROS middleware, known as a node, is responsible for a specific task. Nodes can transmit or receive data from other nodes using a publish/subscribe model utilizing messages passed through logical channels called topics. XTENTH-CAR utilizes ROS to communicate sensor information to the computation unit for processing, and transmit computed control signals to the actuators.

The original ROS, now referred to as ROS 1, launched in 2007 with its last distribution release being ROS Noetic Ninjemys, built for Ubuntu 20.04 (Focal Fossa) release. It will be supported until May 2025 \cite{ros2022noetic}. The new version of ROS, ROS 2, is a substantial revision of the ROS Application-Program Interface (API) designed to take advantage of contemporary technologies and libraries, and provide embedded hardware and real-time code support \cite{macenski2022ros2}. XTENTH-CAR software is written for both ROS 1 Noetic Ninjemys and ROS 2 Foxy Fitzroy to take advantage of the mature ROS 1 ecosystem, while ensuring continued support and improvements to ROS via ROS 2.

\section{Actuation} \label{actuation}

\begin{figure}[ht]
    \centering
    \includegraphics[width = 0.8\columnwidth]{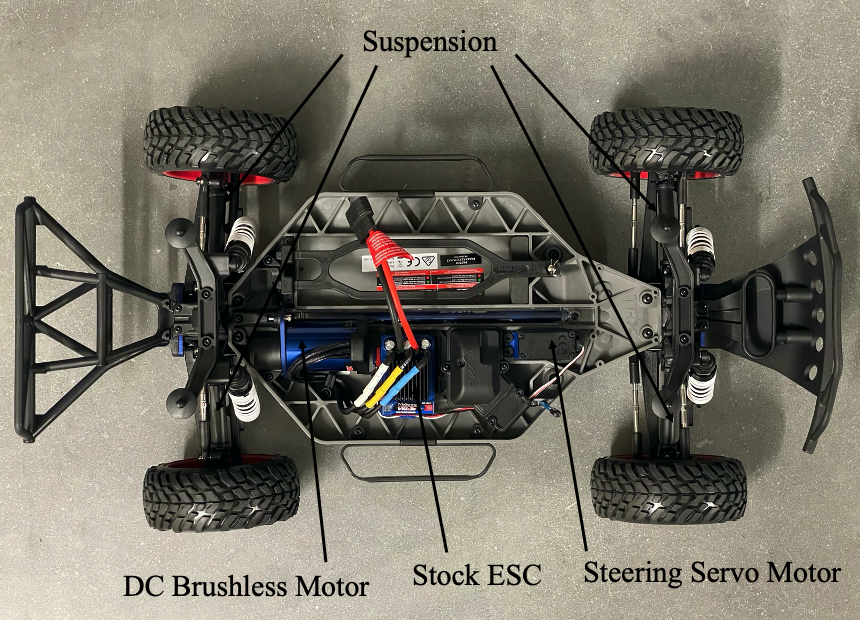}
    \caption{Proportionally scaled chassis used for XTENTH-CAR development.} \label{chassis}
\end{figure}

In order to realistically emulate full-scale CAVs, a scaled platform must accurately simulate on-road vehicle dynamics. XTENTH-CAR is built around a proportionally 1/10 scale Traxxas Slash 4X4 VXL all-terrain Remote Controlled (RC) Ackermann-driven chassis \cite{traxxas}. This detailed chassis, depicted in Figure \ref{chassis}, features all-terrain long-travel suspension with oil-filled shock absorbers, heavy-duty driveshafts, slipper clutch for traction control and tunable steel gear differentials.

\begin{figure}[ht]
    \centering
    \includegraphics[width = 0.68\columnwidth]{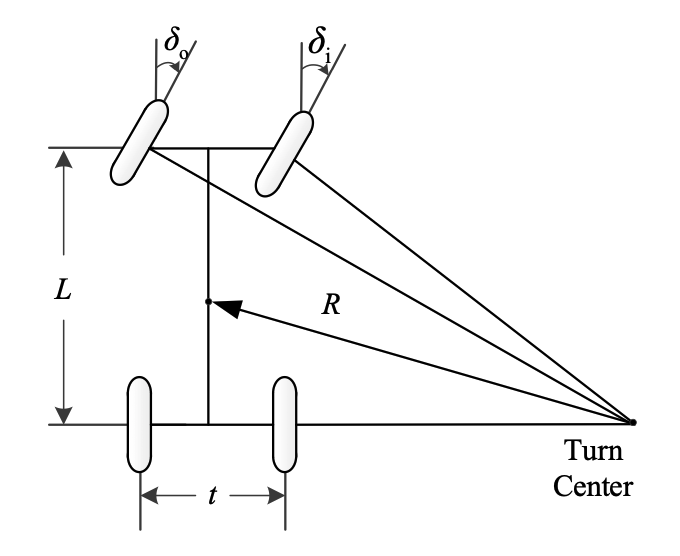}
    \caption{Ackermann steering geometry \cite{gillespie1992fundamentals}.} \label{Ackermann}
\end{figure}

The 10-turn, 3500 $kV$ brushless motor used for propulsion enables a top speed of 60 $mph$, and a high-torque servo motor facilitates Ackermann steering. Figure \ref{Ackermann} portrays the geometry of Ackermann steering, where $\delta_{i}$ and $\delta_{o}$ are steering angles of the inside and outside wheels respectively, $L$ is the wheelbase, $t$ is the axle track and $R$ is the turning radius \cite{gillespie1992fundamentals}.

\noindent $\delta_{i}$ and $\delta_{o}$ are derived as follows,

\begin{equation} \label{}
\delta_{i} = tan^{-1}\left(\frac{L}{R-\frac{t}{2}}\right)
\end{equation}
\begin{equation} \label{}
\delta_{o} = tan^{-1}\left(\frac{L}{R+\frac{t}{2}}\right)
\end{equation}

%The kinematic model of a car-like vehicle is represented by the following state equations \cite{weinstein2010pose},

%\begin{equation} \label{}
%\dot{x} = \textit{v}\,cos(\frac{\pi}{2} - \theta) 
%\end{equation}\
%\begin{equation} \label{}
%\dot{y} = \textit{v}\,sin(\frac{\pi}{2} - \theta)
%\end{equation}\
%\begin{equation} \label{}
%\dot{\theta} = \frac{\textit{v}}{L}\,tan(\delta) 
%\end{equation}\

%Here $x$ and $y$ coordinates describe the vehicle's position, $\theta$ is the bearing, $v$ is the vehicle velocity and $\delta$ is the steering angle of a virtual central front wheel.

\begin{figure}[ht]
    \centering
    \includegraphics[angle=-90,scale=0.12]{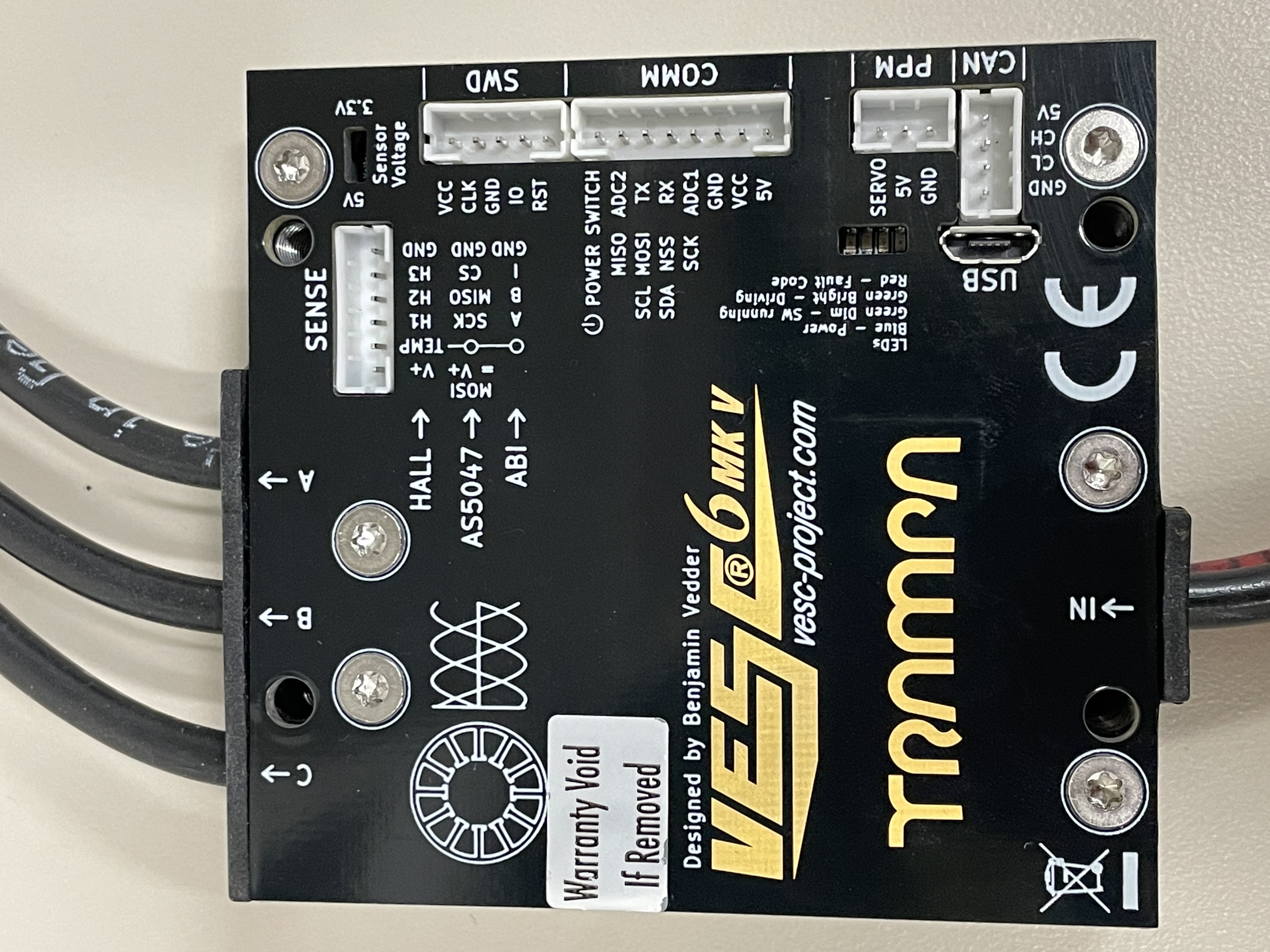}
    \caption{VESC open-source Electronic Speed Controller.} \label{vesc}
\end{figure}

The two actuators are controlled using a VESC, an open-source ESC which replaces the stock ESC included with the chassis. The VESC, depicted in Figure \ref{vesc}, converts velocity and steering angle commands into motor Pulse Width Modulation (PWM) signals, and relays odometry for localization and motion planning via ROS. A 5000 $mAh$ 11.1 $V$ 3-Cell LiPo Battery is utilized to power the VESC and actuators.

The Ackermann control inputs, $v$ and $\delta$ are converted to DC brushless motor Revolutions per Minute (RPM), $V_{m}$ and servo motor position, $\phi_{s}$ as follows,

\begin{equation} \label{}
V_{m} = K_{m}v + m_{o}
\end{equation}
\begin{equation} \label{}
\phi_{s} = K_{s}\delta + s_{o}
\end{equation}

Here the tunable parameters $K_{m}$ and $m_{o}$ are vehicle velocity to motor RPM gain and offset, and $K_{s}$ and $s_{o}$ are steering angle to servo position gain and offset respectively.

The chassis shares similar structure to a full-size on-road vehicle, and is governed by the same physics. The chassis specifications are listed in Table \ref{chassisSpecs}. 

\begin{table}[ht]
    \renewcommand{\arraystretch}{1.4}
    \centering
    \caption{Chassis Specifications}
    \label{chassisSpecs}
        \begin{tabular}{c c}
            \hline
            \textbf{Parameter} & \textbf{Length (inches)} \\
            \hline
            Total Length & 23.36 \\
            \hline
            Axle Track & 11.65 \\
            \hline
            Wheelbase & 12.75 \\
            \hline
            Center Ground Clearance & 2.83 \\
            \hline
            Tire Diameter & 4.31 \\
            \hline
            Inner Wheel Diameter & 3.0 \\
            \hline
            Outer Wheel Diameter & 2.2 \\
            \hline
        \end{tabular}
\end{table}

Tires are interchangeable, and can be swapped for different applications. The stock tires included with the chassis are suitable for indoor CAV experiments. Traxxas canyon trail tires offer greater traction in outdoor environments, thus are a good addition for all-terrain AGV research. 

\section{Computation and Data Processing} \label{computation}

\begin{figure}[ht]
    \centering
    \includegraphics[scale = 0.14]{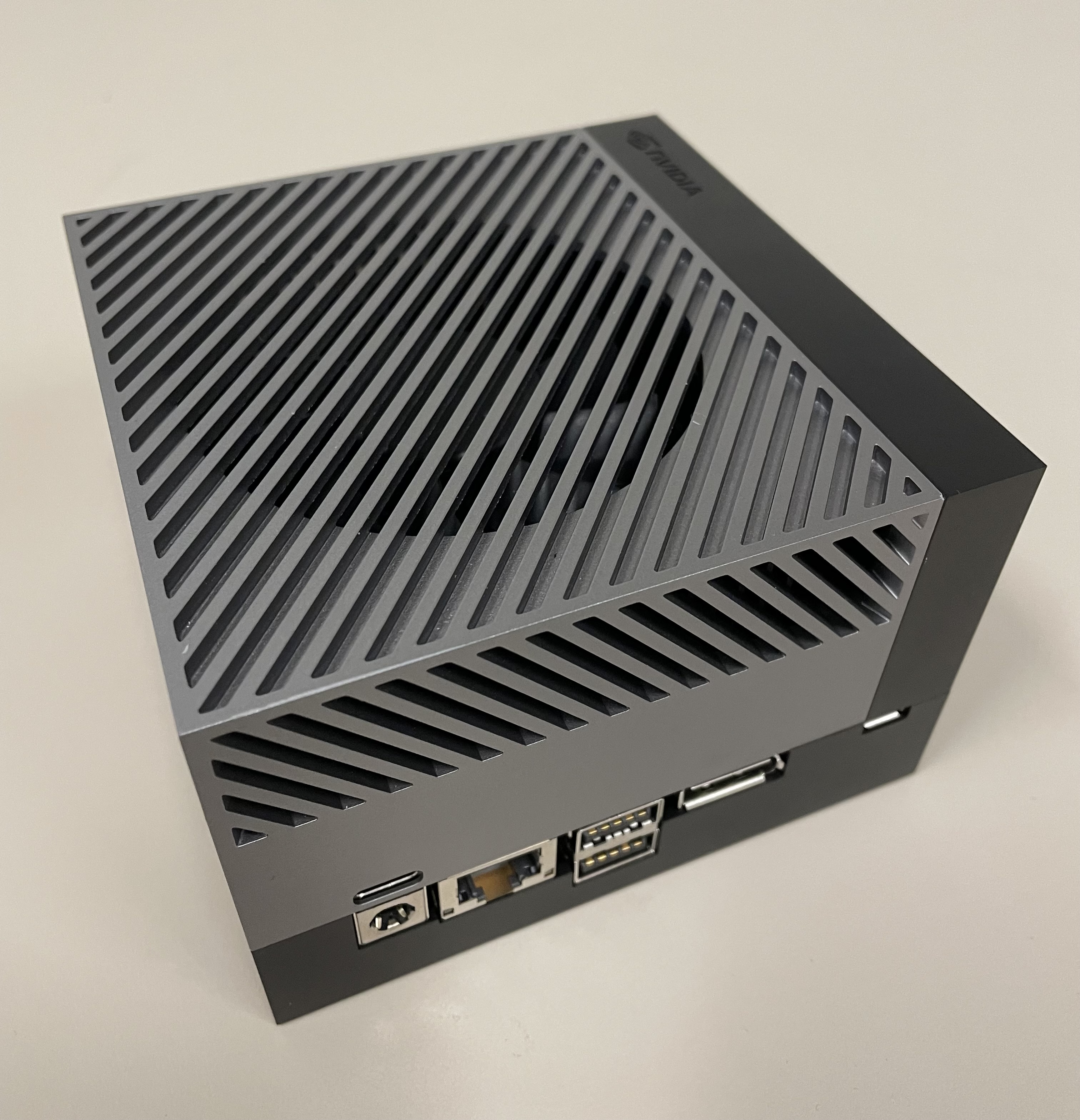}
    \caption{NVIDIA Jetson AGX Orin Developer Kit.} \label{jetsonOrin}
\end{figure}

The NVIDIA Jetson AGX Orin Developer Kit, illustrated in Figure \ref{jetsonOrin}, provides XTENTH-CAR's computation and data processing requirements. This SOM is NVIDIA's newest, and most powerful, capable of 275 TOPS AI performance and up to eight times greater performance than the previous generation Jetson AGX Xavier. The performance upgrade facilitates real-world DRL research, and high-speed experiments that require fast on-board processing.

The small form factor developer kit carries high speed interfaces that include two USB Type-C connectors, four USB Type-A connectors, USB Micro-B connector, DisplayPort, microSD slot and M.2 Key M slot with x4 PCIe Gen4 to facilitate sensor connectivity and expandable storage. The SOM components are listed in Table \ref{SOMcomponents}.

\begin{table}[ht]
    \renewcommand{\arraystretch}{1.4}
    \centering
    \caption{SOM Components}
    \label{SOMcomponents}
        \begin{tabular}{c c}
            \hline
            \textbf{Item} & \textbf{Component} \\
            \hline
            CPU & 12-core Arm® Cortex®-A78AE \\ & v8.2 64-bit \\
            \hline
            GPU & 2048-core NVIDIA Ampere \\ & 64 Tensor Cores \\
            \hline
            Memory & 32GB 256-bit LPDDR5 \\
            \hline
            Boot Drive & WD\_BLACK 1TB SN770 \\ & M.2-2280 NVMe SSD \\
            \hline
            Internal Storage & 64GB eMMC 5.1 \\
            \hline
        \end{tabular}
\end{table}

\begin{figure}[!h]
    \centering
    \includegraphics[scale=0.03]{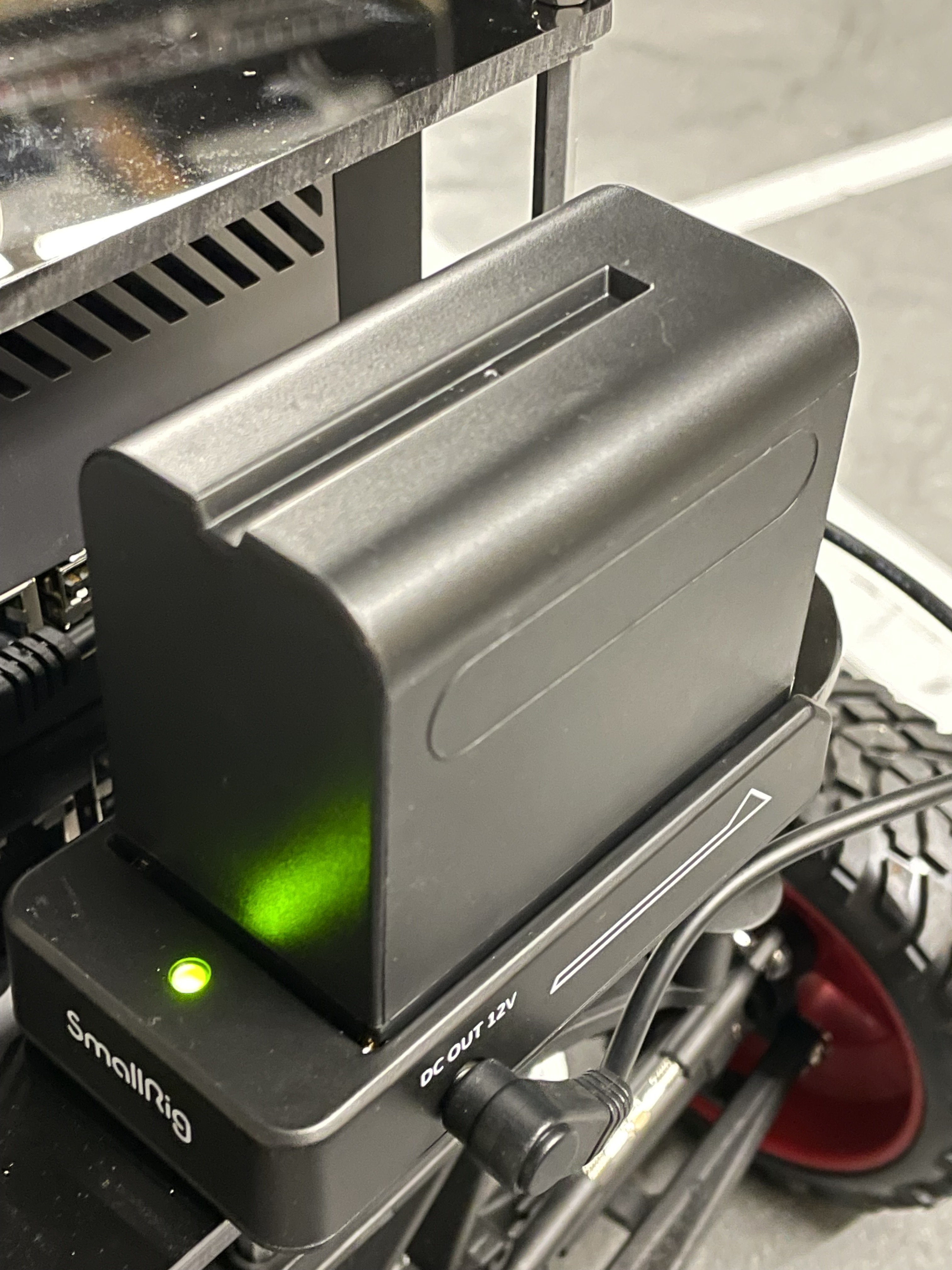}
    \caption{NP-F970 Battery and SmallRig NP-F Adapter.} \label{npfbattery}
\end{figure}

\begin{figure}[!h]
    \centering
    \includegraphics[width = 0.85\columnwidth]{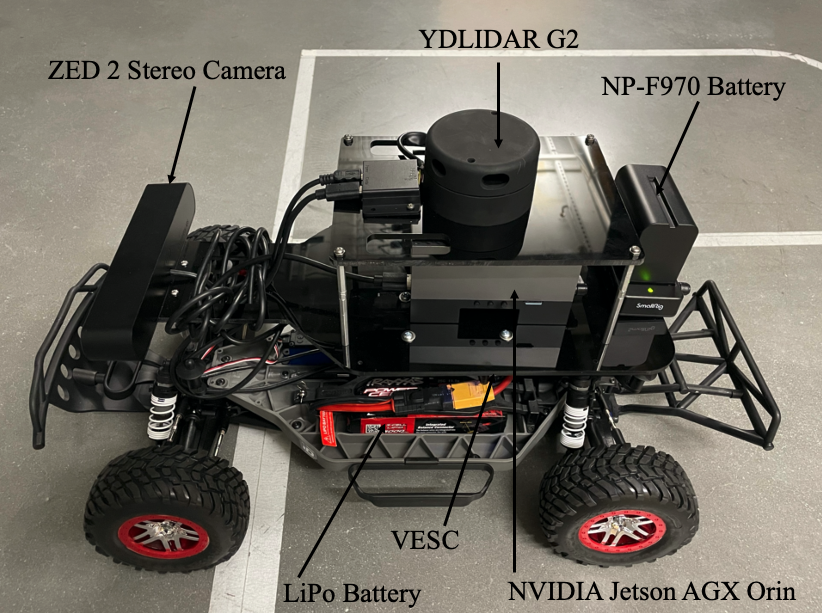}
    \caption{Fully assembled XTENTH-CAR.} \label{xtenthcar}
\end{figure}

The SOM and connected sensors are powered utilizing a 8800 $mAh$ NP-F970 Battery mounted on a SmallRig NP-F Battery Adapter Plate, as illustrated in Figure \ref{npfbattery}. All electronics are mounted on custom acrylic laser cut platforms as shown in Figure \ref{xtenthcar}. The use of separate batteries to power the SOC and actuators facilitates long lasting experiments.

\section{Sensing} \label{sensing}

The XTENTH-CAR platform consists of a 120{\textdegree} FoV stereo camera with built-in IMU, barometer and magnetometer, and a 360{\textdegree} 2D LiDAR. Pose and velocity are accurately determined by the VESC, and Wi-Fi is used as a communication interface to interchange information.

\subsection{Camera}

\begin{figure}[!h]
    \centering
    \includegraphics[scale=0.15]{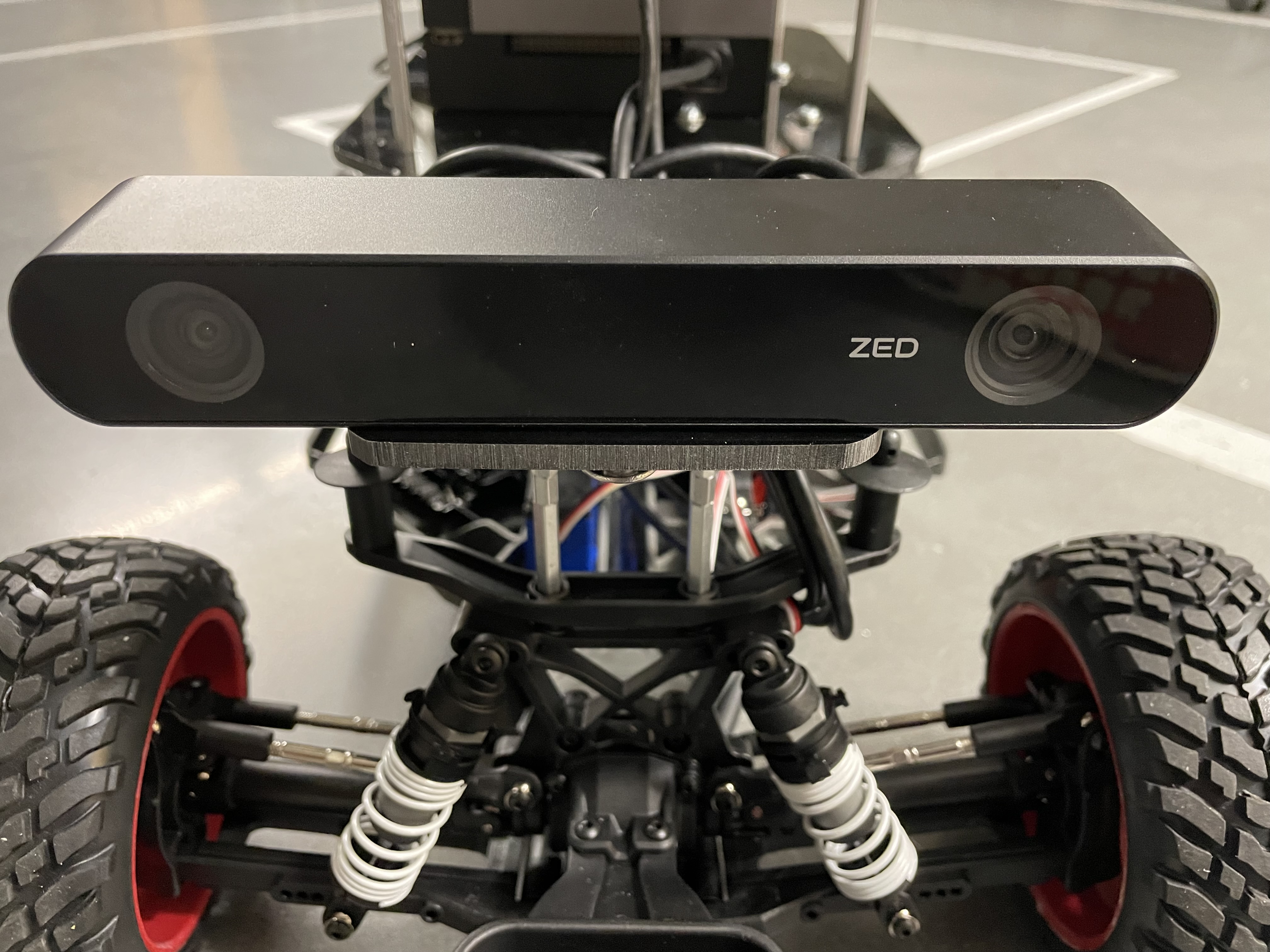}
    \caption{ZED 2 stereo camera mounted at the front of the chassis.} \label{zed2}
\end{figure}

Cameras capture light from the surrounding environment to create images, and convert them into electric signals for processing. Images can be used to detect road surfaces, lanes, traffic signs and lights, and objects. Sequential camera images enable dynamic object tracking, and visual odometry for pose estimation \cite{soleimanitaleb2019object, Nister2004visual}. We use a ZED 2 stereo camera from Stereolabs, illustrated in Figure \ref{zed2}, for CV in XTENTH-CAR. The similarly specced, rugged, dust, water and humidity resistant ZED 2i is suitable for outdoor experiments.

The camera system contains two camera sensors that each have an output resolution of 2208x1242 at 15 Frames per Second (fps), 1920x1080 at 30 fps, 1280x720 at 60 fps and 672x376 at 100 fps, with a combined FoV of 120{\textdegree} and depth range of 20 $m$ to simulate human binocular vision, and capture 3D images. Moreover, ZED 2 includes an IMU, a barometer and a magnetometer for pose estimation, and atmospheric pressure and magnetic field measurements. The camera specifications are listed in Table \ref{ZED2Specs}. 

\begin{table}[ht]
    \renewcommand{\arraystretch}{1.4}
    \centering
    \caption{ZED 2 Stereo Camera Specifications}
    \label{ZED2Specs}
        \begin{tabular}{c c}
            \hline
            \textbf{Characteristic} & \textbf{Specification} \\
            \hline
            Max. Output Resolution & 2208x1242 at 15 fps \\
            \hline
            Min. Output Resolution & 672x376 at 100 fps \\
            \hline
            Field of View & 120{\textdegree} \\
            \hline
            Depth Range & 0.3-20 $m$ \\
            \hline
            Depth Accuracy & $<$1\% up to 3 $m$, \\ & $<$5\% up to 15 $m$  \\
            \hline
            Sensors & IMU, Barometer \\ & and Magnetometer \\
            \hline
        \end{tabular}
\end{table}

%ZED Software Development Kit (SDK) includes modules for spatial object detection and tracking, and neural depth sensing to facilitate ML CV.

\subsection{LiDAR}

\begin{figure}[ht]
    \centering
    \includegraphics[scale=0.033]{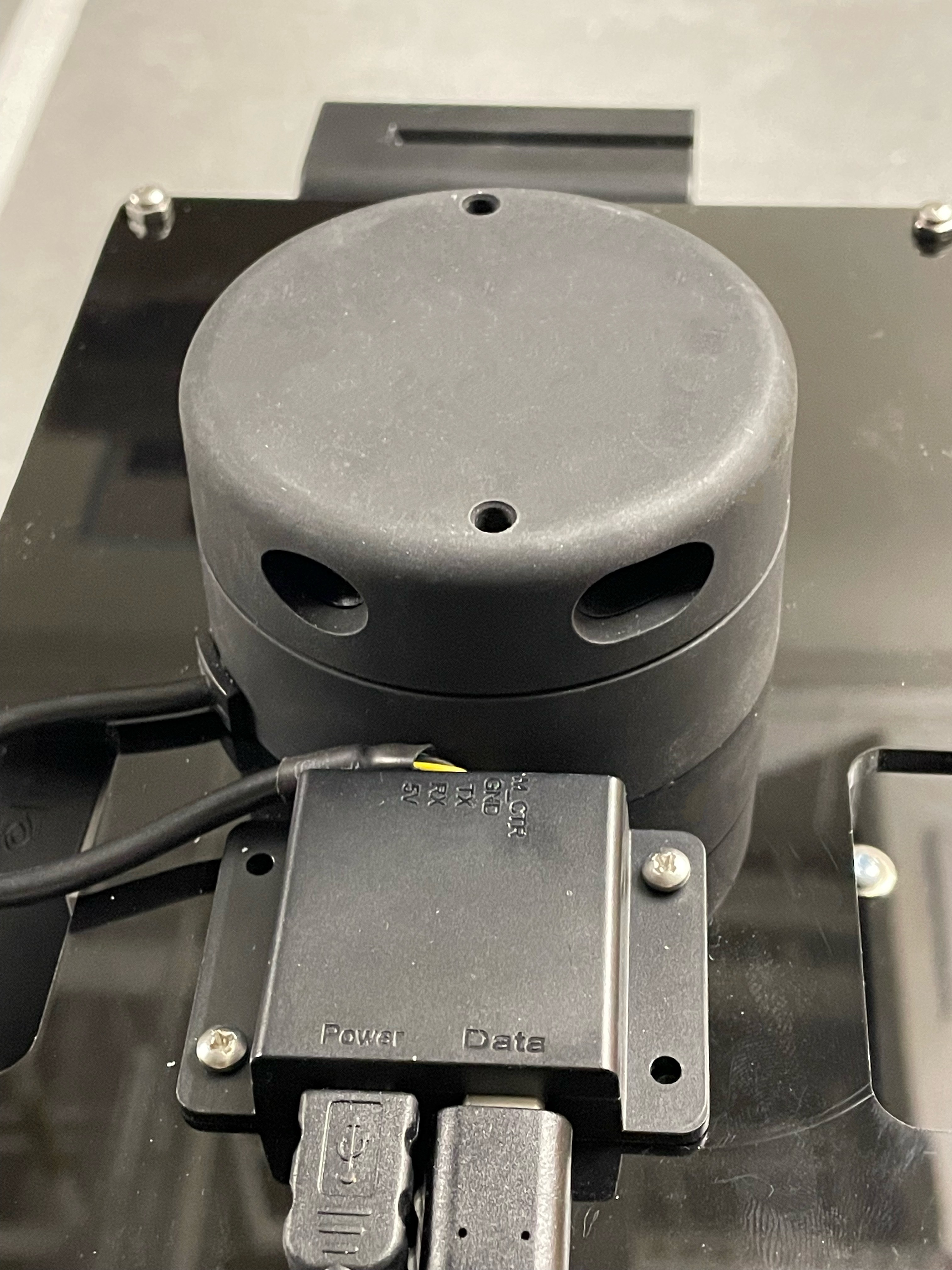}
    \caption{YDLIDAR G2 mounted on a level platform.} \label{ydlidarG2}
\end{figure}

 LiDAR (Light Detection and Ranging) is an active, remote sensing method that creates a precise point cloud map of the surrounding environment by emitting pulsed laser beams and measuring the time taken for reflected light to return to the system \cite{raj2020survey}. The sensor can be used to detect and determine distances of objects from the ego vehicle. Full-size AVs utilize one or more 3D LiDAR sensors for mapping and localization. We use a YDLIDAR G2, portrayed in Figure \ref{ydlidarG2}, for XTENTH-CAR, a 2D 360{\textdegree} LiDAR with a maximum range of 12 $m$. The LiDAR specifications are listed in Table \ref{lidarSpecs}.

\begin{table}[ht]
    \renewcommand{\arraystretch}{1.4}
    \centering
    \caption{YDLIDAR G2 Specifications}
    \label{lidarSpecs}
        \begin{tabular}{c c}
            \hline
            \textbf{Characteristic} & \textbf{Specification} \\
            \hline
            Scan Angle & 360{\textdegree} \\
            \hline
            Scan Frequency & 5-12 $Hz$ \\
            \hline
            Range Distance &  0.12-12 $m$ \\
            \hline
            Range Frequency & 5000 $Hz$ \\
            \hline
            Angle Resolution & 0.36{\textdegree}-0.864{\textdegree}  \\
            \hline
        \end{tabular}
\end{table}

The LiDAR is mounted 11.5 inches from the surface of the ground, thus provides a 2D planar Point Cloud Map (PCM) of the surrounding environment at this height. This works especially well for CAV and AGV experiments on a small-scale platform, since experiments can accommodate the sensor's capabilities. 

3D 360{\textdegree} LiDARs such as the Velodyne Puck draw more power, require substantially greater computation resources, and are heavier, which can lead to the small-scale chassis bottoming out. YDLIDAR G2 is lightweight, and has low computation and power consumption requirements which make it ideal for a scaled experimental vehicle platform. 

\subsection{Communication}

CAVs such as X-CAR use Dedicated Short-Range Communication (DSRC) for direct inter-vehicle, and vehicle to infrastructure communication \cite{kenney2011dedicated}. These one-way or two-way channels are specifically designed for automobiles in the Intelligent Transportation System (ITS) and a corresponding set of protocols and standards, eliminating reliance on cellular equipment and related infrastructure.

XTENTH-CAR utilizes either Transmission Control Protocol/Internet Protocol (TCP/IP) or User Datagram Protocol (UDP) for data transfer via Wi-Fi. The NVIDIA Jetson AGX Orin Developer Kit includes embedded Wi-Fi, and does not require a separate Wi-Fi card that is needed in previous versions.

\section{Results and Discussion} \label{results}

In this section we discuss XTENTH-CAR's real-world performance, and capabilities.

\subsection{Environmental Perception}

\subsubsection{Mapping and Localization}

\begin{figure}[!h]
    \centering
    \includegraphics[width = 0.8\columnwidth]{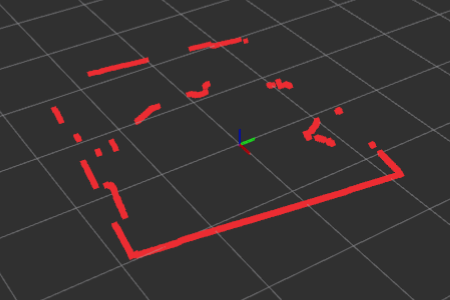}
    \caption{2D point cloud map generated by YDLIDAR G2 with obstacles and surfaces marked in red.} \label{2Dpcm}
\end{figure}

The 2D PCM generated by YDLIDAR G2 is depicted in Figure \ref{2Dpcm}, with obstacles and surfaces marked in red. The vehicle's pose is conveyed by the coordinate frame. 

\begin{figure}[!h]
    \centering
    \includegraphics[width = 0.4\columnwidth]{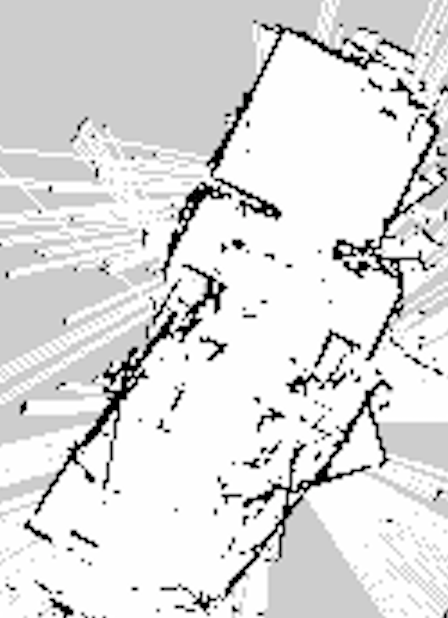}
    \caption{Map of an office space at Virginia Tech generated using Hector SLAM.} \label{slam}
\end{figure}

%XTENTH-CAR maps its surrounding environment, localizes and detects obstacles in a 2D plane utilizing LiDAR PCM. 
2D LiDAR SLAM is computationally less expensive than visual SLAM \cite{filipenko2018comparison}, which allocates more resources for demanding real-time motion planning and control algorithms such as DRL and Nonlinear Model Predictive Control (NMPC) \cite{sivashangaran2022racing}. Figure \ref{slam} illustrates a map of an office space at Virginia Tech generated using Hector SLAM.

\subsubsection{Object Detection, Tracking and Identification}

\begin{figure}[ht]
    \centering
    \includegraphics[width = \columnwidth]{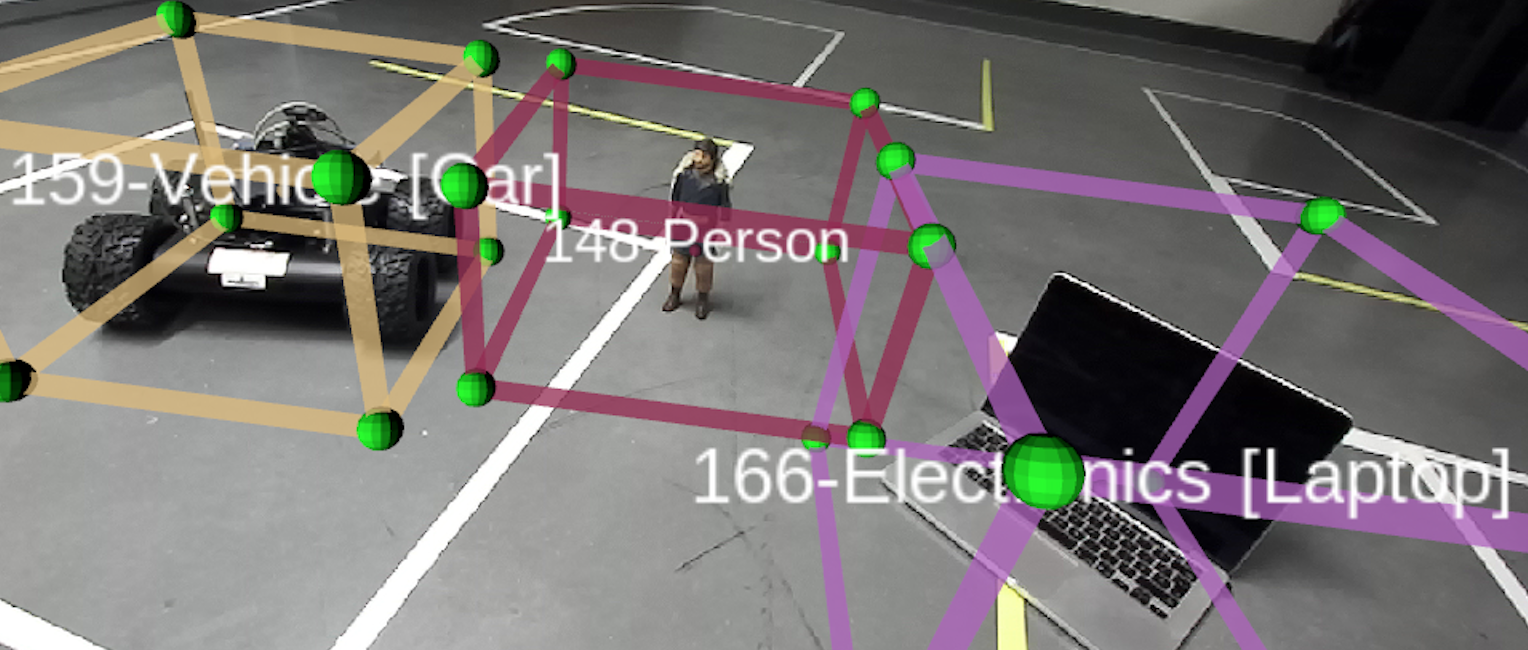}
    \caption{Object detection and tracking utilizing Mask R-CNN.} \label{objDectCamera}
\end{figure}

Camera images and Mask R-CNN were utilized for real-time object detection, tracking and identification as shown in Figure \ref{objDectCamera}.

The CV ML algorithm accurately identifies a laptop, a humanoid action figure labelled as a person and a four-wheel differential drive mobile robot labelled as a vehicle. Similar RC humanoid action figures and wheeled mobile robots can emulate pedestrians and road traffic in 1/10th scaled CAV experiments. 

\subsection{Actuator Control}

\begin{figure}[ht]
    \centering
    \includegraphics[width = \columnwidth]{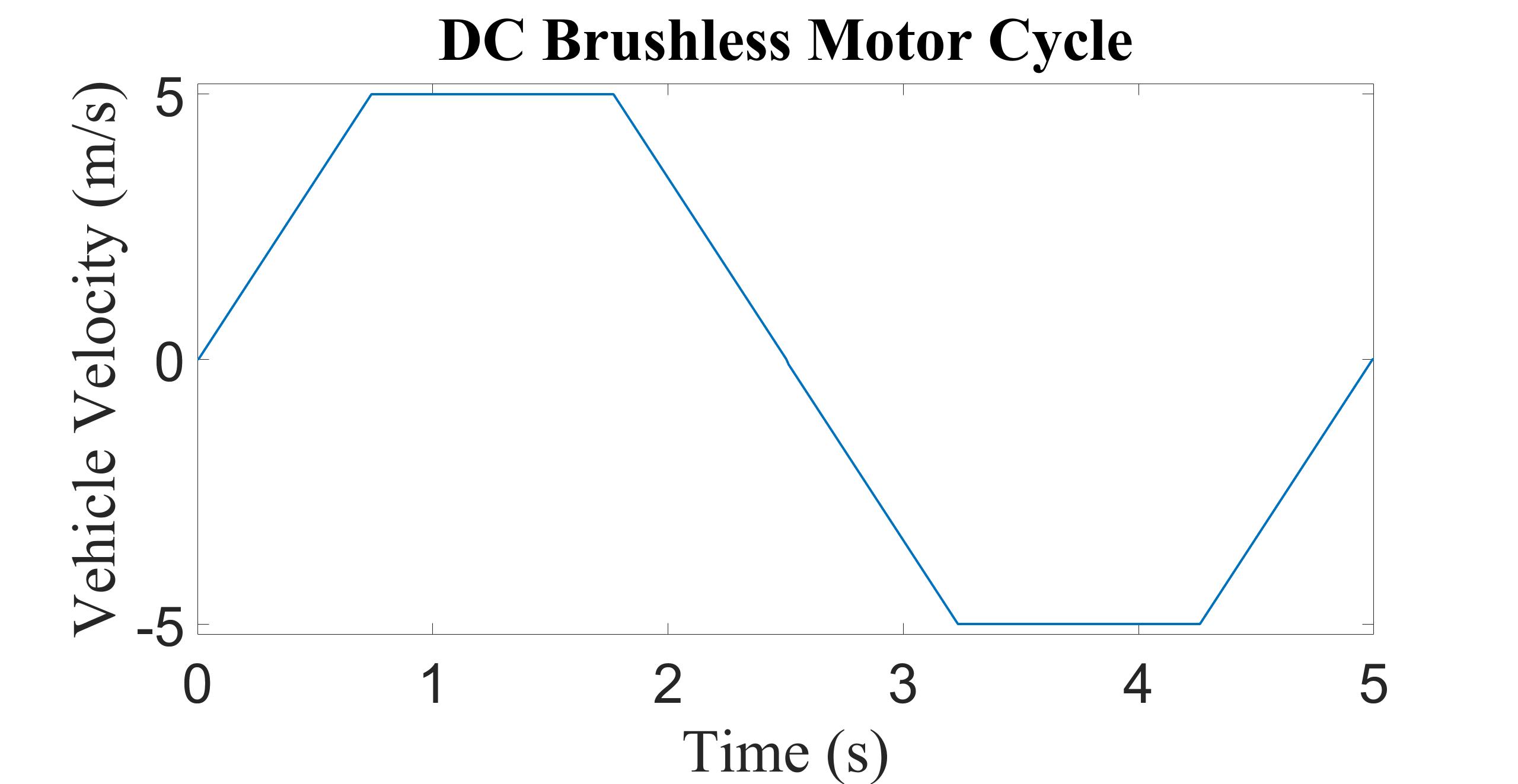}
    \setlength{\abovecaptionskip}{-\baselineskip}
    \caption{DC brushless motor cycle for full range of vehicle velocity control inputs.} \label{dcMotorCycle}
\end{figure}

The two actuators, DC brushless motor and steering servo motor, are controlled by the VESC efficiently with minimal noise. Figure \ref{dcMotorCycle} illustrates the DC brushless motor cycle for a full range of velocity control inputs.

Maximum and minimum velocities are set to 5 and -5 $m/s$ by default for safety in indoor environments, but can be configured to be higher. The DC brushless motor accelerates XTENTH-CAR from 0 to 5 $m/s$ in 0.75 $s$ with an acceleration of 6.67 $m/s^{2}$. Deceleration performance is equivalent with the vehicle decelerating from 5 to -5 $m/s$ in 1.5 $s$ at -6.67 $m/s^{2}$.   

\begin{figure}[ht]
    \centering
    \includegraphics[width = \columnwidth]{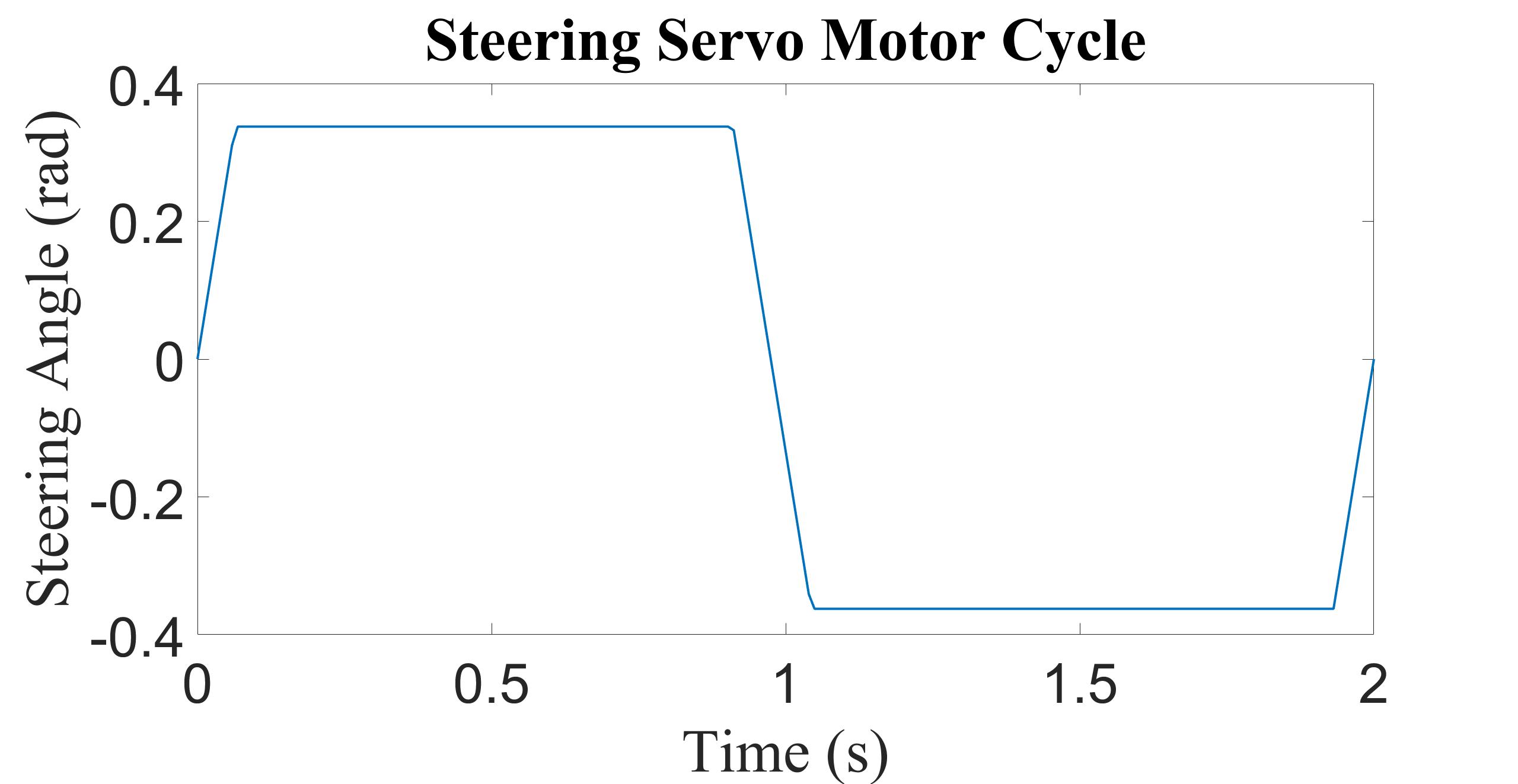}
    \setlength{\abovecaptionskip}{-\baselineskip}
    \caption{Steering servo motor cycle for full range of vehicle steering angle control inputs.} \label{steeringServoCycle}
\end{figure}

The steering servo motor cycle for a full range of steering angle control inputs is illustrated in Figure \ref{steeringServoCycle}.

The maximum and minimum steering angles are $0.36^{c}$ and $-0.36^{c}$ respectively. The steering servo motor directs the front wheels to the maximum steering angle in either direction from $0^{c}$ in 0.069 $s$ at a rate of 5.22 $rad/s$.

\subsection{Computation Performance}

A range of autonomous tasks were tested to evaluate real-time computation performance. The CPU and memory usage percentages for actuator control, LiDAR PCM, Hector SLAM, CV object detection and tracking, collision avoidance using MPC integrated with APF \cite{shang2023mpcapf}, and cognitive exploration utilizing DRL \cite{sivashangaran2023deep}, trained for simulation to real-world policy transfer in AutoVRL (AUTOnomous ground Vehicle deep Reinforcement Learning simulator) \cite{sivashangaran2023autovrl}, a high-fidelity AGV simulator we developed using open-source tools for sim-to-real DRL research and development, are listed in Table \ref{computation_table}.

\begin{table}[ht]
    \renewcommand{\arraystretch}{1.4}
    \centering
    \caption{CPU and Memory Usage for Various Autonomous Tasks}
    \label{computation_table}
        \begin{tabular}{c c c}
            \hline
            \textbf{Task} & \textbf{CPU Usage \%} & \textbf{Memory Usage \%} \\
            \hline
            Actuator Control & 6.4 & 8.8 \\
            \hline
            Actuator Control\\ \& PCM & 7.0 & 9.0 \\
            \hline
            Actuator Control, \\ PCM \& SLAM & 14.0 & 11.7 \\
            \hline
            Actuator Control, \\ PCM \& MPC & 15.7 & 11.7 \\
            \hline
            Actuator Control, \\ PCM \& DRL & 7.6 & 12.6 \\
            \hline
            Actuator Control, \\ PCM \& CV & 23.3 & 15.7 \\
            \hline
            Actuator Control, \\ PCM, CV \& MPC & 28.2 & 16.1 \\
            \hline
            Actuator Control, \\ PCM, CV \& DRL & 24.2 & 17.0 \\
            \hline
        \end{tabular}
\end{table}

CPU usage ranges for these tasks from 6.4\% to 28.2 \%, and memory usage from 8.8 \% to 17.0 \%. MPC is more CPU demanding than DRL, as it performs more computations at each time step when compared to a trained DRL policy. Conversely, the DRL agent requires more memory to load the trained policy. CV using Mask R-CNN was the most demanding task, nonetheless XTENTH-CAR handles the full perception load, motion planning with MPC or DRL, and actuator control efficiently, utilizing a fraction of the available CPU and memory resources enabling improved real-time performance and high sampling times for high speed experiments.

\section{Conclusions} \label{conclusions}

This paper introduced XTENTH-CAR, a proportionally scaled experimental vehicle platform for connected autonomy and all-terrain research with open-source software stack written for both ROS 1 \& ROS 2. The platform utilizes the NVIDIA Jetson AGX Orin Developer Kit for best-in-class computation and data processing to facilitate experimental CAV and AGV research in state-of-the-art computationally expensive domains such as Deep Reinforcement Learning (DRL). After providing background, we discussed actuation to emulate a full-scale on-road CAV on a proportionally scaled platform. Moreover, we detailed the embedded computation unit, and sensing suite utilized to produce a complete understanding of the scaled vehicle's surrounding environment, required for experimental evaluation of CAV and AGV motion planning and control research. A range of autonomous tasks that include LiDAR PCM, SLAM and ML based CV for perception, MPC and DRL for motion planning, and actuator control were tested in real-time to evaluate computation performance, and shown to execute efficiently, utilizing a maximum of 28.2 \% CPU and 17.0 \% memory usage enabling high sampling times, and improved real-time performance.

%%%%%%%%%%%%%%%%%%%%%%%%%%%%%%%%%%%%%%%%%%%%%%%%%%%%%%%%%%%%%%%%%%%%%%
%\begin{acknowledgment}
%ASME Technical Publications provided the format specifications for the Journal of Mechanical Design, though they are not easy to reproduce.  It is their commitment to ensuring quality figures in every issue of JMD that motivates this effort to have authors review the presentation of their figures.  

%Thanks go to D. E. Knuth and L. Lamport for developing the wonderful word processing software packages \TeX\ and \LaTeX. We would like to thank Ken Sprott, Kirk van Katwyk, and Matt Campbell for fixing bugs in the ASME style file \verb+asme2ej.cls+, and Geoff Shiflett for creating 
%ASME bibliography stype file \verb+asmems4.bst+.
%\end{acknowledgment}

%%%%%%%%%%%%%%%%%%%%%%%%%%%%%%%%%%%%%%%%%%%%%%%%%%%%%%%%%%%%%%%%%%%%%%
% The bibliography is stored in an external database file
% in the BibTeX format (file_name.bib).  The bibliography is
% created by the following command and it will appear in this
% position in the document. You may, of course, create your
% own bibliography by using thebibliography environment as in
%
% \begin{thebibliography}{12}
% ...
% \bibitem{itemreference} D. E. Knudsen.
% {\em 1966 World Bnus Almanac.}
% {Permafrost Press, Novosibirsk.}
% ...
% \end{thebibliography}

% Here's where you specify the bibliography style file.
% The full file name for the bibliography style file 
% used for an ASME paper is asmems4.bst.
\bibliographystyle{asmems4}

% Here's where you specify the bibliography database file.
% The full file name of the bibliography database for this
% article is asme2e.bib. The name for your database is up
% to you.
\bibliography{asme2ej}

%%%%%%%%%%%%%%%%%%%%%%%%%%%%%%%%%%%%%%%%%%%%%%%%%%%%%%%%%%%%%%%%%%%%%%
%\appendix       %%% starting appendix
%\section*{Appendix A: Head of First Appendix}
%Avoid Appendices if possible.

%%%%%%%%%%%%%%%%%%%%%%%%%%%%%%%%%%%%%%%%%%%%%%%%%%%%%%%%%%%%%%%%%%%%%%
%\section*{Appendix B: Head of Second Appendix}
%\subsection*{Subsection head in appendix}
%The equation counter is not reset in an appendix and the numbers will
%follow one continual sequence from the beginning of the article to the very end as shown in the following example.
%\begin{equation}
%a = b + c.
%\end{equation}

\end{document}